\documentclass[10pt,twocolumn,letterpaper]{article}

\usepackage{cvpr}              

\usepackage{cite}
\usepackage{amsmath,amssymb,amsfonts}
\usepackage{graphicx}
\usepackage{textcomp}
\usepackage{mathtools}
\usepackage{multirow}
\usepackage{booktabs}
\usepackage{subcaption}
\usepackage{algorithm}
\usepackage{algorithmicx}
\usepackage{algpseudocode}
\usepackage{appendix}
\usepackage[pagebackref,breaklinks,colorlinks]{hyperref}

\graphicspath{ {fig/} }
\usepackage[utf8]{inputenc}
\usepackage[capitalize]{cleveref}
\crefname{section}{Sec.}{Secs.}
\Crefname{section}{Section}{Sections}
\Crefname{table}{Table}{Tables}
\crefname{table}{Tab.}{Tabs.}

\def\cvprPaperID{*****} 
\def\confName{CVPR}
\def\confYear{2022}

\begin{document}

\title{Latent Denoising Diffusion GAN: Faster sampling, Higher image quality}

\author{Luan Thanh Trinh, Tomoki Hamagami \\
Yokohama National University\\
{\tt\small tt.luan.ynu@gmail.com, hamagami@ynu.ac.jp}
}
\maketitle

\begin{abstract}
Diffusion models are emerging as powerful solutions for generating high-fidelity and diverse images, often surpassing GANs under many circumstances. However, their slow inference speed hinders their potential for real-time applications. To address this, DiffusionGAN leveraged a conditional GAN to drastically reduce the denoising steps and speed up inference. Its advancement, Wavelet Diffusion, further accelerated the process by converting data into wavelet space, thus enhancing efficiency. Nonetheless, these models still fall short of GANs in terms of speed and image quality. To bridge these gaps, this paper introduces the Latent Denoising Diffusion GAN, which employs pre-trained autoencoders to compress images into a compact latent space, significantly improving inference speed and image quality. Furthermore, we propose a Weighted Learning strategy to enhance diversity and image quality. Experimental results on the CIFAR-10, CelebA-HQ, and LSUN-Church datasets prove that our model achieves state-of-the-art running speed among diffusion models. Compared to its predecessors, DiffusionGAN and Wavelet Diffusion, our model shows remarkable improvements in all evaluation metrics. Code and pre-trained checkpoints: \url{https://github.com/thanhluantrinh/LDDGAN.git}
\end{abstract}


\section{Introduction}
\label{sec:introduction}
Despite being a recent introduction, diffusion models have quickly established themselves as a pivotal paradigm for image-generation tasks. At their core, diffusion models hinge on two crucial processes: the forward process (or diffusion process) and the reverse process (denoising process). In the forward process, Gaussian noise is incrementally infused into the input image until it transforms into an isotropic Gaussian. Conversely, in the reverse process, a model is meticulously trained to invert the forward process and faithfully reproduce the original input image. After training, the power of diffusion models shines as we can generate high quality images by navigating randomly sampled noise through the adeptly learned denoising process.

In comparison to other prominent deep generative models, diffusion models distinguish themselves through the excellence of the generated images in terms of quality and diversity, coupled with their inherent training stability. Particularly noteworthy is the observation that diffusion models have surpassed Generative Adversarial Networks (GANs), which have dominated the image generation task in recent years, excelling in both image quality (\cite{fid1, fid2}) and diversity (\cite{recall1, recall2, Likelihood SDE}). One notable aspect heightening expectations for diffusion models is their growing capacity to effectively incorporate various conditional inputs, such as semantic maps, text, representations, and images. This versatility expands the potential applications of diffusion models into areas like text-to-image generation (\cite{ld, fid2, text2img}), video generation (\cite{video1, video2}), image-to-image translation (\cite{img2img1, img2img2, img2img3}), text-to-3D generation \cite{3D}, and beyond.

Despite their considerable potential, the hindrance of slow inference speed poses a significant obstacle for diffusion models in becoming fully-fledged image generation models that can meet diverse expectations across various domains, particularly in real-time applications. The fundamental cause of slow sampling in these models lies in the Gaussian assumption made during the denoising step, an assumption that is valid only for small step sizes. Consequently, diffusion models often necessitate a substantial number of denoising steps, typically ranging in the hundreds or thousands. By modeling complex and multimodal distributions through conditional GANs, DiffusionGAN \cite{ddgan} enables larger denoising steps, reducing the number of denoising steps to just a few, thereby significantly accelerating the inference time of diffusion models. Wavelet Diffusion \cite{wddgan}, an enhancement of DiffusionGAN, achieves a further increase in inference speed by transferring data from pixel space to wavelet space, reducing the input data size by a factor of 4 and becoming the fastest existing diffusion model. However, Wavelet Diffusion still lags considerably behind StyleGAN \cite{StyleGAN2}. Additionally, the acceleration of inference speed shows signs of compromising the output image quality, as the output quality of both DiffusionGAN and Wavelet Diffusion is lower than that of StyleGAN and other recent diffusion models.

This paper aims to bridge both the gap in image quality and the gap in speed by introducing the Latent Denoising Diffusion GAN (LDDGAN). Firstly, instead of residing in a high-dimensional pixel space, input images are compressed as much as possible into a low-dimensional latent space through pre-trained autoencoders. This compression significantly reduces computational costs during both training and inference, facilitating faster sampling. Given that the latent space is more suitable for diffusion models than the high-dimensional pixel space \cite{ld}, our approach aims to enhance both image quality and sample diversity by utilizing this space for both the diffusion and denoising processes. Following the principles of DiffusionGAN, a conditional GAN is employed to model complex and multimodal distributions and enable a large denoising step. Additionally, to enhance diversity through adversarial loss while leveraging the effect of reconstruction loss for image quality improvement, we propose a novel learning strategy called Weighted Learning. Since our model primarily uses multimodal distributions, instead of restricting latents learned by the autoencoder to Gaussian distributions as in other latent-based models (\cite{ld, vae, Score SDE}), we allow the autoencoder to freely search for the appropriate latent space. This approach helps to significantly accelerate the convergence of the autoencoder and improve the overall quality and diversity of the main model.

Based on experimental results on standard benchmarks including CIFAR-10, CELEBA-HQ, and LSUN Church, our model achieves state-of-the-art running speed for a diffusion model while maintaining high image quality. When compared to GANs, we achieve comparable image generation quality and running speed, along with increased diversity. Additionally, in comparison to two predecessors, DiffusionGAN and Wavelet Diffusion, our model significantly outperforms them across the majority of comparison metrics.

In summary, our contributions are as follows:
\begin{itemize}
    \item We propose a novel Latent Denoising Diffusion GAN framework that leverages dimensionality reduction and the high compatibility of low-dimensional latent spaces with the diffusion model's denoising process. This approach not only improves inference speed and the quality of generated images but also enhances diversity.
    \item We find that if the denoising process of a diffusion model does not depend on Gaussian distributions, it becomes necessary to eliminate the autoencoder's learned dependency on the Gaussian distribution to enhance the diversity and quality of the generated images. This insight could serve as a recommendation for future latent-based diffusion models.
    \item We propose an innovative Weighted Learning strategy that boosts diversity through adversarial loss, while also improving image quality via the effect of reconstruction loss.
    \item Our Latent Denoising Diffusion GAN features low training costs and state-of-the-art inference speeds, paving the way for real-time, high-fidelity diffusion models.
\end{itemize}

\section{Related Work}
\label{sec: related work}
\subsection{Image Generation models}
GANs (Generative Adversarial Networks, \cite{gan}) are among the representative generative models extensively utilized in various real-time applications due to their ability to rapidly generate high-quality images \cite{3 of ld}. However, challenges in optimization (\cite{2 ld, 28 ld, 54 ld}) and difficulty in capturing the full data distribution \cite{55 ld} represent two main weaknesses of GANs. In contrast, Variational Autoencoders (VAEs, \cite{vae}) and flow-based models (\cite{flow1, flow2}) excel in fast inference speed and high sample diversity but face challenges in generating high-quality images. Recently, diffusion models have been introduced and have quickly made an impressive impact by achieving state-of-the-art results in both sample diversity \cite{recall2} and sample quality \cite{15 ld}. One of the two primary weaknesses of diffusion models, high training costs, has been effectively addressed by Latent Diffusion Models (LDMs, \cite{ld}). This was achieved by shifting the diffusion process from the pixel space to the latent space using pre-trained autoencoders. Despite the similarity between this approach and ours in using latent space, their reliance on Gaussian distributions in both the denoising process and the latent space learned by the autoencoder necessitates thousands of network evaluations during the sampling process. This dependency results in slow inference speeds, representing the primary remaining weakness of diffusion models. Consequently, it becomes a significant bottleneck, restricting the practical use of diffusion models in real-time applications.

\subsection{Faster Diffusion Models}
To enhance the inference speed of diffusion models, several methods have been proposed, including learning an adaptive noise schedule \cite{rel 1}, using non-Markovian diffusion processes \cite{DDPM, FastDDPM}, and employing improved SDE solvers for continuous-time models \cite{rel 2}. However, these methods either experience notable deterioration in sample quality or still necessitate a considerable number of sampling steps. Studies on knowledge distillation \cite{DDPM Distillation, distillation1, distillation2} are also worth noting, as in some cases, they can perform the diffusion process with just a few steps or even just one step and produce high-quality images. The weakness of the knowledge distillation method is that it requires a well-pre-trained diffusion model to be used as a teacher model for the main model (student model). The results of the student model often struggle to surpass the teacher model due to this student-teacher constraint.

The most successful method to date for accelerating inference speed without compromising image quality, and without relying on another pre-trained diffusion model, is DiffusionGAN \cite{ddgan}. By modeling complex and multimodal distributions through conditional GANs, DiffusionGAN enables larger denoising steps, reducing the number of denoising steps to just a few and significantly accelerating the inference process. Wavelet Diffusion \cite{wddgan}, an enhancement of DiffusionGAN, achieves a further increase in inference speed by transferring data from pixel space to wavelet space, reducing the input data size by a factor of 4 and becoming the fastest existing diffusion model. However, the inference speed of Wavelet Diffusion is still much slower than that of traditional GANs, and the trade-off between sampling speed and quality still requires further improvement. We aim to address both problems by utilizing latent space.

\subsection{Latent-based approaches}
Regarding diffusion model families, similar to our approach, LDMs in \cite{ld} and Score-based generative models (SGMs) in \cite{Score SDE} utilize an autoencoder to transform data from pixel space to latent space and conduct the diffusion process in this space. However, as mentioned above, their dependence on Gaussian distributions in both the denoising process and the latent space learned by the autoencoder results in the necessity for thousands of network evaluations during the sampling process. This dependency leads to slow inference speeds. Our method enables larger denoising steps, thereby reducing the number of denoising steps to just a few and significantly accelerating the inference speed. Additionally, due to the absence of a reliance on Gaussian distribution, unlike LDMs and SGMs, our autoencoders are free to explore potential latent spaces rather than attempting to model spaces with Gaussian distribution. Details about our autoencoders are described in Section~\ref{subsec:autoencoder}.

\section{Background}
\label{sec:background}
\subsection*{Diffusion Models}
The main idea of Diffusion models (\cite{ld,DDPM,Score SDE,Likelihood SDE}) lies in gradually adding noise to the data and then training a model to reverse that process gradually. In this way, the model becomes capable of generating data similar to the input data from pure Gaussian noise. The forward process, in which we gradually add noise to the input data $\textbf{x}_0$, can be defined as follows:

\begin{equation}
    \begin{matrix}
    q(\textbf{x}_{1:T}|\textbf{x}_{0}) = \prod_{t \geq 1} q(\textbf{x}_{t}|\textbf{x}_{t-1}),   \\
    with \quad q(\textbf{x}_{t}|\textbf{x}_{t-1}) = \mathcal{N} (\textbf{x}_{t}; \sqrt{1-\beta_t}, \beta_t \textbf{I})
\end{matrix}
\label{eq: forward}
\end{equation}
where $T$ and $\beta_t$ denote number of steps and pre-defined variance schedule at timesteps $t$, respectively. The reverse process, in which we remove noise from data, can be defined: 

\begin{equation}
    \begin{matrix}
    p_{\theta}(\textbf{x}_{0:T}) = p(\textbf{x}_T)\prod_{t \geq 1}   p_{\theta}(\textbf{x}_{t-1}|\textbf{x}_{t}),   \\
    with \quad p_{\theta}(\textbf{x}_{t-1}|\textbf{x}_{t}) = \mathcal{N} (\textbf{x}_{t-1}; \boldsymbol{\mu}_{\theta}(\textbf{x}_t,t), \sigma_t^2 \textbf{I})
\end{matrix}
\label{eq: backward}
\end{equation}
where $\boldsymbol{\mu}_{\theta}(\textbf{x}_t,t)$ and $\sigma_t^2$ are the mean and variance for the denoising model parameterized by $\theta$. As introduced in \cite{b4}, the optimal mean can be defined as follows.

\begin{equation}
    \boldsymbol{\mu}(\textbf{x}_t,t) = \frac{1}{\sqrt{\alpha_t}}\Big( \textbf{x}_{t} - \frac{\beta_t}{\sqrt{1- \overline{\alpha}_t }} \mathbb{E} [\boldsymbol{\epsilon}|\textbf{x}_t]\Big)
\end{equation}
Here, $\overline{\alpha}_t = \prod_{i=1}^t \alpha_i$, $\alpha_t = 1 - \beta_t$, and $\epsilon$ is the noise added to $\mathbf{x_t}$. Therefore, the learning objective can be considered as a noise prediction task.

\begin{equation}
    \min_{\theta} \mathbb{E}_{t, \textbf{x}_0, c, \boldsymbol{\epsilon}} \left\| \boldsymbol{\epsilon} - \boldsymbol{\epsilon}_{\boldsymbol{\theta}}(\textbf{x}_t, t, c) \right\|^2_2
    \label{eq:original_loss}
\end{equation}

with $c$ is the condition fed into the model (e.g., class-conditional \cite{Score SDE} or text- to-image \cite{ld}). Eq. \ref{eq: backward} suggests that diffusion models often make the assumption that the denoising distribution can be approximated by Gaussian distributions. Nevertheless, it is widely acknowledged that the Gaussian assumption only holds in the infinitesimal limit of small denoising steps (as highlighted by Sohl-Dickstein \textit{et al.} in \cite{b2} and Feller in \cite{b3}). This necessitates a large number of steps in the reverse process and causes a slow sampling issue for diffusion models.

\subsection*{Denoising Diffusion GAN and Wavelet Diffusion}
\label{subsec: DDGAN}
To reduce the inference time of diffusion models, it's crucial to significantly decrease the number of denoising diffusion steps $T$ required for the reverse process. Additionally, a new denoising method is required because using large denoising steps causes the denoising distributions $p_{\theta}(\textbf{x}_{t-1}|\textbf{x}_{t})$ to become more complex and multimodal, unlike the existing diffusion models (\cite{ld,DDPM,Score SDE}) where the denoising distributions follow a Gaussian distribution. To overcome this challenge, DiffusionGAN models the denoising distribution with an complex and multimodal distribution by using conditional GANs. Whereas existing diffusion models predict noise $\boldsymbol{\epsilon}$ added to $x_{t-1}$ at timestep $t$ using $x_{t}$, the generator $G_{\theta}(x_t, z, t)$ in DiffusionGAN predicts the input data $x_0$ with random latent variable $z \sim\mathcal{N}(0, \textbf{I})$. This crucial distinction allows DiffusionGAN's denoising distribution $p_{\theta}(\textbf{x}_{t-1}|\textbf{x}_t)$ to become multimodal and complex, unlike the unimodal denoising model of existing approaches. The perturbed sample $x'_{t-1}$ is then acquired using pre-defined $q(x_{t-1}|x_t, x_0)$. The discriminator $D_{\phi}$ performs judgment on fake pairts $D_{\phi}(x'_{t-1}, x_t, t)$ and real pairs $D_{\phi}(x_{t-1}, x_t, t)$.

Building upon Diffusion GAN, Wavelet Diffusion further speeds up inference by performing denoising in wavelet space rather than pixel space. It first decomposes the input image $x \in \mathbb{R}^{3 \times H \times W}$ into its low-frequency and high-frequency components using a wavelet transformation and concatenates them to form a matrix $y \in \mathbb{R}^{12 \times \frac{H}{2} \times \frac{W}{2}}$. This helps to reduce the input data size by 4 times and decrease the inference time of the model as well as the computational cost. To better exploit the features and improve the output image quality, Wavelet Diffusion proposes a wavelet-embedded generator optimized for wavelet space. Wavelet Diffusion also introduces a reconstruction loss between the generated data and its ground truth to boost the model training convergence.

In the following sections, DDGAN and WDDGAN will stand for DiffusionGAN and Wavelet Diffusion, respectively, for brevity.

\section{Method}
\label{sec:method}
In this section, we begin by providing a comprehensive overview of the proposed framework, Latent Denoising Diffusion GAN (Section~\ref{subsec:overview}). Following this, we introduce the autoencoder utilized to train the main model (Section~\ref{subsec:autoencoder}). Finally, we discuss the effect of reconstruction loss on the final result and introduce a novel training strategy called Weighted Learning (Section~\ref{subsec:loss functions}).
\begin{figure}[h]
    \centering
    \includegraphics[width=0.5\textwidth]{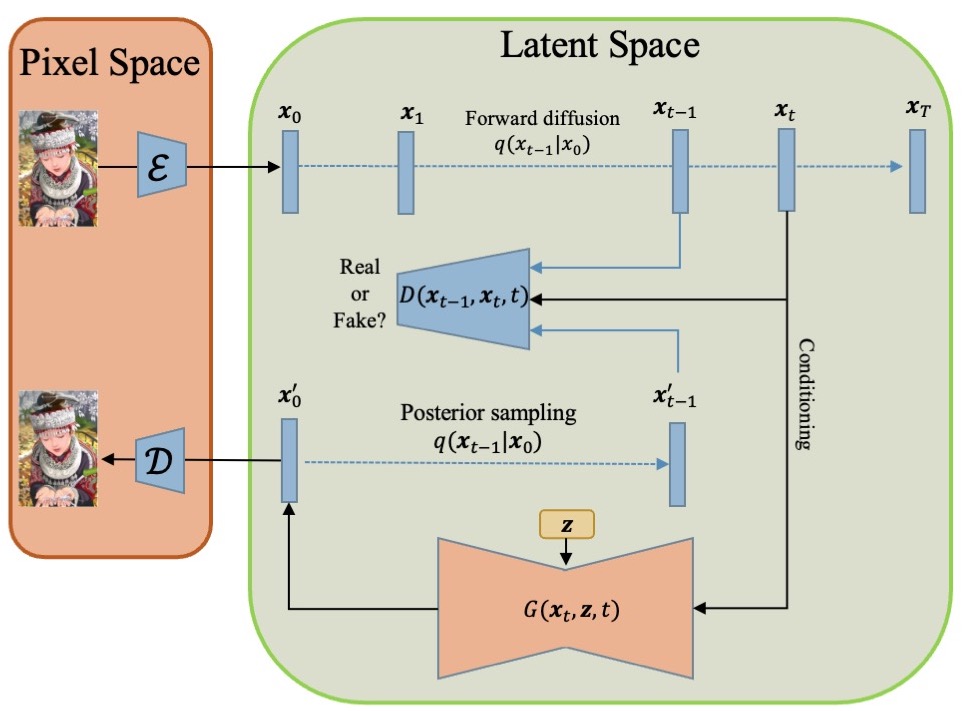}
    \caption{The training process of Latent Denoising Diffusion GAN (LDDGAN)}
    \label{fig:overview}
\end{figure}

\subsection{Latent Denoising Diffusion GAN}
\label{subsec:overview}
Figure~\ref{fig:overview} provides an overview of our proposed LDDGAN. A pre-trained encoder is used to compress input data $X_0$ and transform it from pixel space to low-dimensional latent representation $x_0 = \mathcal{E}(X_0)$. It is important to note that the encoder downsamples the image by a factor of $f$. We strive to compress the data as much as possible (using a larger $f$), while still guaranteeing image quality upon decoding. Details about the autoencoder are described in Section~\ref{subsec:autoencoder}.

The forward diffusion process and the reverse process are performed in low-dimensional latent space, rather than in wavelet space as in WDDGAN, or pixel space as in DDGAN and other inherent diffusion models. This method has two main advantages. Firstly, LDDGAN is capable of reducing the input image size by factors of 4, 8, or even more, unlike WDDGAN which can only reduce it by a factor of 4. This enhancement significantly increases the model's inference speed and reduces computational costs. As investigated by Rombach \textit{et al.} \cite{ld}, compared to high-dimensional pixel space, low-dimensional latent space is better suited for the diffusion model, a likelihood-based generative method. This results in improved quality and diversity of the generated samples.

In the latent space, to reduce inference time, the forward diffusion process is performed with a small number of sampling steps $T$ $(T \leq 8)$, and each diffusion step has a significantly larger $\beta_t$ compared to traditional diffusion models. At timestep $t$, a corrupted sample $\textbf{x}_t$ is generated from the input sample $\textbf{x}_0$ using the posterior distribution $q(\textbf{x}_{t-1}|\textbf{x}_0)$. The generator $G(\textbf{x}_t, \textbf{z}, t)$ uses $\textbf{x}_t$, the current timestep $t$, and a random latent variable $\textbf{z} \sim \mathcal{N}(0, \textbf{I})$ to predict an approximation of the original sample $\textbf{x}'_0$. The predicted sample $\textbf{x}'_t$ is then calculated using the same posterior distribution $q(\textbf{x}_{t-1}|\textbf{x}_0)$. Finally, the discriminator evaluates both the fake pairs $D(\textbf{x}'_{t-1}, \textbf{x}_t, t)$ and real pairs $D(\textbf{x}_{t-1}, \textbf{x}_t, t)$ to provide feedback to the generator. The decoder $\mathcal{D}$ is not necessary for training but is used to convert $\textbf{x}'_0$ back to the output image. We describe the sampling process in Algorithm~\ref{alg:LDDGAN}.

\begin{algorithm}[H]
\caption{Sampling process of LDDGAN}\label{alg:LDDGAN}
\begin{algorithmic}[1]
\State $\textbf{x}_T \sim \mathcal{N}(0, \textbf{I})$
\For{$t = T, \ldots, 1, 0$}
    \State $\textbf{z} \sim \mathcal{N}(0, \textbf{I})$
    \State $\textbf{x}'_0 = G(\textbf{x}_t, \textbf{z}, t)$
    \State $\textbf{x}'_{t-1} \sim q(\textbf{x}_{t-1}|\textbf{x}_0)$
\EndFor
\State $X_0 = \mathcal{D}(\textbf{x}'_0)$
\Return $X_0$
\end{algorithmic}
\end{algorithm}

\subsection{Autoencoder}
\label{subsec:autoencoder}
In the context of latent space representation learning, recent studies, particularly diffusion models (\cite{ld}, \cite{ld2}, \cite{ld3}) and variational autoencoders (VAEs, \cite{VAE1, VAE2}), frequently employ a KL-penalty (Kullback-Leibler divergence, \cite{KL, KL2}) between the Gaussian distribution and the learned latent within the loss function. This approach encourages the learned latent space to closely approximate a Gaussian distribution, proving effective when model assumptions rely heavily on Gaussian distributions. However, with models that allow for complex and multimodal distributions, as in this study, we do not use this KL-penalty. Instead, we allow our autoencoder to freely explore latent spaces and prioritize the ability to compress and recover images effectively.

We construct our autoencoder architecture based on VQGAN \cite{vqgan}, distinguished by the integration of a quantization layer within the decoder. To ensure adherence to the image manifold and promote local realism in reconstructions, we employ a training regimen that encompasses both perceptual loss \cite{106-ld} and a patch-based adversarial loss \cite{23-ld, 103-ld}, similar to VQGAN. This approach effectively circumvents the potential for blurriness that can arise when training relies exclusively on pixel-space losses such as L2 or L1 objectives. Furthermore, motivated by investigations in \cite{ld} that have demonstrated the superiority of two-dimensional latent variables over traditional one-dimensional counterparts in the domains of image compression and reconstruction, we elected to utilize a two-dimensional latent variable as both the output of our encoder and the input to the subsequent diffusion process.

\subsection{Loss Functions}
\label{subsec:loss functions}
\subsubsection{Adversarial loss}
The adversarial loss for the discriminator and the generator are defined as follows.

\begin{equation}
    \mathcal{L}^{D}_{adv} = - \log(D(x_{t-1}, x_t, t)) + \log(D(x'_{t-1}, x_t, t))
\end{equation}
\begin{equation}
    \mathcal{L}^{G}_{adv} = - \log(D(x'_{t-1}, x_t, t))
\end{equation}

\subsubsection{Reconstruction loss and Weighted Learning}
\label{subsec: WL}
In WDDGAN \cite{wddgan}, in addition to the adversarial loss, a reconstruction loss between the generated sample and its ground truth is also employed during the training of the generator. The generator’s overall objective function is mathematically formulated as a linear combination of adversarial loss and reconstruction loss, weighted by a fixed hyperparameter. It is expressed as follows:

\begin{equation}
    \mathcal{L}^{G}_{rec} = ||\textbf{x}_0 - \textbf{x}'_0||
\end{equation}
\begin{equation}
    \mathcal{L}_{G} = \mathcal{L}^{D}_{adv} + \lambda \mathcal{L}^{G}_{rec}
    \label{eq:original loss G}
\end{equation}

As investigated in \cite{wddgan}, using reconstruction loss helps WDDGAN improve image fidelity. This enhancement can be attributed to the fact that when only adversarial loss is used, the generator learns to produce $\textbf{x}'_0$ that is similar to the reference input $\textbf{x}_0$ indirectly, through feedback from the discriminator. However, reconstruction loss provides the generator with direct feedback through the similarity between $\textbf{x}_0$ and $\textbf{x}'_0$, calculated using L1 loss. Therefore, employing reconstruction loss in conjunction with adversarial loss facilitates the training of the generator easier and achieve better convergence. There is also evidence in \cite{wddgan} that WDDGAN converges faster than DDGAN. However, we postulate that simply employing a linear combination of the two loss functions, as presented in Eq.~\ref{eq:original loss G}, to construct the overall loss function proves ineffective and potentially reduces the diversity of generated images. This hypothesis stems from two primary considerations. Firstly, at a certain training stage, the generator acquires the ability to produce data closely resembling the input data solely through the guidance of the discriminator's feedback, effectively negating the necessity of reconstruction loss. Secondly, reconstruction loss may cause the generator to tend to generate data that is identical to the training data for any given input noise, thereby restricting its capacity for creative exploration and diverse generation.

\begin{figure}[h]
    \centering
    \includegraphics[width=0.5\textwidth]{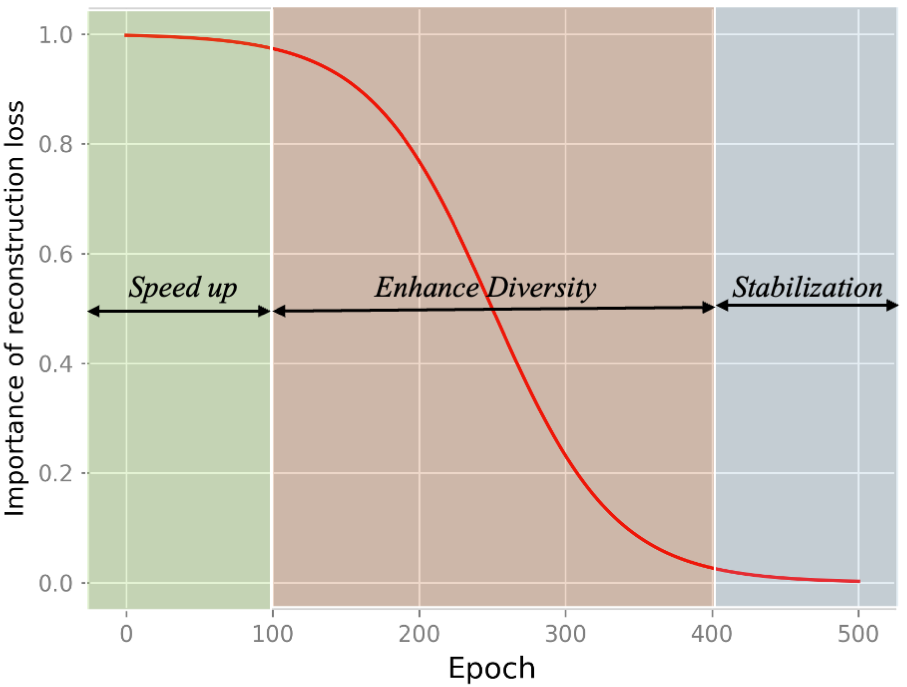}
    \caption{An example of Weighted Learning}
    \label{fig:Weighted Learning}
\end{figure}

In this study, we also apply reconstruction loss. However, instead of using a simple linear combination, we propose a method called Weighted Learning to combine the overall loss function of the generator, which is detailed as follows:

\begin{equation}
    \begin{matrix}
    \mathcal{L}_{G} = \mathcal{L}^{D}_{adv} + \lambda \mathcal{L}^{G}_{rec} \\
    \\
    with \quad \varphi = - \delta + \delta \frac{Epoch}{NumEpoch} \quad and \quad \lambda = 1 - \frac{1}{1+\exp^{-\varphi}}

\end{matrix}
\label{eq: Weighted Learning}
\end{equation}
Here, $\delta$ is a weighting hyper-parameter and Figure~\ref{fig:Weighted Learning} is an example of our method. In the early stages of training, the importance of reconstruction loss remains largely unchanged with $\lambda \simeq 1$. This configuration helps to leverage reconstruction loss to mitigate the weaknesses of adversarial loss, which arises from indirect learning through the discriminator, and increases the convergence speed of the model. In the subsequent stage, the importance of reconstruction loss is reduced, giving priority to adversarial loss. This adjustment is expected to help increase the diversity of generated images. Finally, as training nears completion, the importance of reconstruction loss is gradually reduced until it approaches zero, to avoid excessive changes in the overall training objective and enhance training stability.

Before starting with Weighted Learning, we also tried removing the reconstruction function after training the model with both loss functions to a certain extent. However, suddenly changing the training objective caused the model to become unstable and difficult to converge again. The experiments to verify the effectiveness of Weighted Learning are presented in Section~\ref{sec:experiments}.


\section{Experiments}
\label{sec:experiments}
We first present a detailed description of the experimental settings, dataset, and evaluation metrics used in Section~\ref{subsec:setup}. Next, we present experimental results that verify the effectiveness of the proposed LDDGAN and compare these results with previous studies in Section~\ref{subsec:CIFAR10 results}. Finally, we evaluate the effectiveness of Weighted Learning and the impact of the learned latent space by the Autoencoder in Section~\ref{subsec:alab}.
\subsection{Experimental setup}
\label{subsec:setup}
\subsubsection{Datasets}
To save computational costs, we use CIFAR-10 $32\times32$ as the main dataset for qualitative and quantitative comparisons with previous studies, as well as evaluating the effectiveness of each component added to LDDGAN. To further visualize the effectiveness of the model, we use LSUN Church $256\times256$ \cite{lsun} and CELEBA-HQ $256\times256$ \cite{celeba} datasets to assess performance through high-resolution image generation tasks.

\subsubsection{Evaluation metrics}
When evaluating our model, we consider three key factors: inference time, sample fidelity, and sample diversity. To assess inference time, we measure the number of function evaluations (NFE) and the average time taken to generate a batch size of 100 across 300 trials. We use the widely recognized Fréchet Inception Distance (FID, by Heusel \textit{et al.} in \cite{fid}) metric for sample fidelity. To measure sample diversity, we rely on the improved recall score developed by Kynkäänniemi \textit{et al.} in \cite{newRecall}, which is an enhanced version of the original one proposed by Sajjadi \textit{et al.} in \cite{oldRecall}. For comparison, we use FID and Recall data from previously published papers. To ensure that the FID and Recall results remain consistent with those in the original papers, we adopt the code and experimental environment from the published code of the previous studies mentioned above.

\subsubsection{Autoencoder}
As described in Section~\ref{subsec:autoencoder}, we build our autoencoder based on VQGAN \cite{vqgan}, distinguished by the integration of a quantization layer within the decoder. The autoencoder is trained using perceptual loss \cite{106-ld} and patch-based adversarial loss \cite{23-ld, 103-ld} to maintain adherence to the image manifold and promote local realism in the reconstructions. We aim to compress data as much as possible (using a larger scale factor $f$), while still ensuring image quality upon decoding. Table~\ref{tab: autoencoder} lists the autoencoders that were successfully trained for each dataset. All models were trained until convergence, defined as the point where there was no further substantial improvement in FID.
\begin{table}[h!]
    \centering
     \begin{tabular}{l c c c} 
     \hline
     Dataset & Scale factor $f$ & Ouput size & FID \\ 
     \hline
     CIFAR-10  & 2 & $16\times16\times4$ & 1.32\\ 
     CELEBA-HQ  & 4 & $64\times64\times4$ & 0.58 \\ 
     LSUN Church  & 8 & $32\times32\times3$ & 1.14 \\
     \hline
    
     \end{tabular}
     \caption{Successfully trained autoencoders are used for each dataset. The FID is calculated between the reconstructed images after compression and the original images from the validation set. The output size of these autoencoders will serve as the input size for both the discriminator and the generator.}
     \label{tab: autoencoder}
\end{table}

\begin{table*}[h!]
    \centering
     \begin{tabular}{l c c c c c c c c c } 
     \hline
      & \multicolumn{3}{c}{\textbf{DDGAN}} & \multicolumn{3}{c}{\textbf{WDDGAN}} &\multicolumn{3}{c}{\textbf{Ours} (*)} \\
      \cmidrule(lr){2-4}\cmidrule(lr){5-7} \cmidrule(lr){8-10}
      & Params & FLOPs & MEM & Params & FLOPs & MEM & Params & FLOPs & MEM \\
     \hline
     CIFAR-10 (32) & 48.43M & 7.05G & 0.31G & 33.37M &1.67G & 0.16G & 41.97M & 1.72G & 0.18G\\ 
     CELEBA-HQ (256)& 39.73M & 70.82G & 3.21G & 31.48M & 28.54G & 1.07G & 45.39M & 7.15G & 0.27G \\ 
     LSUN(256)& 39.73M & 70.82G & 3.21G & 31.48M & 28.54G & 1.07G &40.92M & 9.68G & 0.25G \\ 
     \hline
    
     \end{tabular}
     \caption{Model specifications of DDGAN \cite{ddgan}, WDDGAN \cite{wddgan} and our approach including a number of parameters (M), FLOPs (GB), and memory usage (GB) on a single GPU for one sample. (*) including the decoder.}
     \label{tab: computational cost}
\end{table*}

\subsubsection{Implementation details}
\label{subsubsec: imple}
Our implementation is based on DDGAN \cite{ddgan}, and we adhered to the same training configurations as those used in DDGAN for our experiments. In constructing our GAN generator, we align with DDGAN's architectural choice by employing the U-Net-structured NCSN++ framework as presented in Song et al. \cite{NCSN}. To effectively model the denoising distribution within a complex and multimodal context, we leverage latent variable $\textbf{z}$ to exert control over normalization layers. To achieve this, we strategically substitute all group normalization layers (Wu and He \cite{groupNorm}) within the generator with adaptive group normalization layers. This technique involves utilizing a straightforward multi-layer fully-connected network to accurately predict the shift and scale parameters within group normalization directly from $\textbf{z}$. The network's input consists of by the conditioning element $\textbf{x}_t$, and time embedding is meticulously employed to ensure conditioning on $t$.

Thanks to the efficient compression of input data through autoencoders, our models require fewer GPU resources compared to WDDGAN and DDGAN. With a batch size equal to or greater than those used by these models, all of our models can be trained on just 1-2 NVIDIA RTX A5000 GPUs (24GB), whereas WDDGAN and DDGAN require 1-8 NVIDIA V100 GPUs (32 GB) or NVIDIA A100 GPUs (40 GB). Our model only requires a configuration similar to DDGAN's CIFAR10 model for high-resolution datasets like LSUN and CelebA-HQ. A detailed comparison of computational costs between DDGAN, WDDGAN, and our models is presented in Table~\ref{tab: computational cost}. Our model has a comparable number of parameters but uses significantly fewer computing FLOPs and less memory.

\subsection{Experimental results}
\label{subsec:CIFAR10 results}

\begin{figure}[h]
    \centering
    \includegraphics[width=0.45\textwidth]{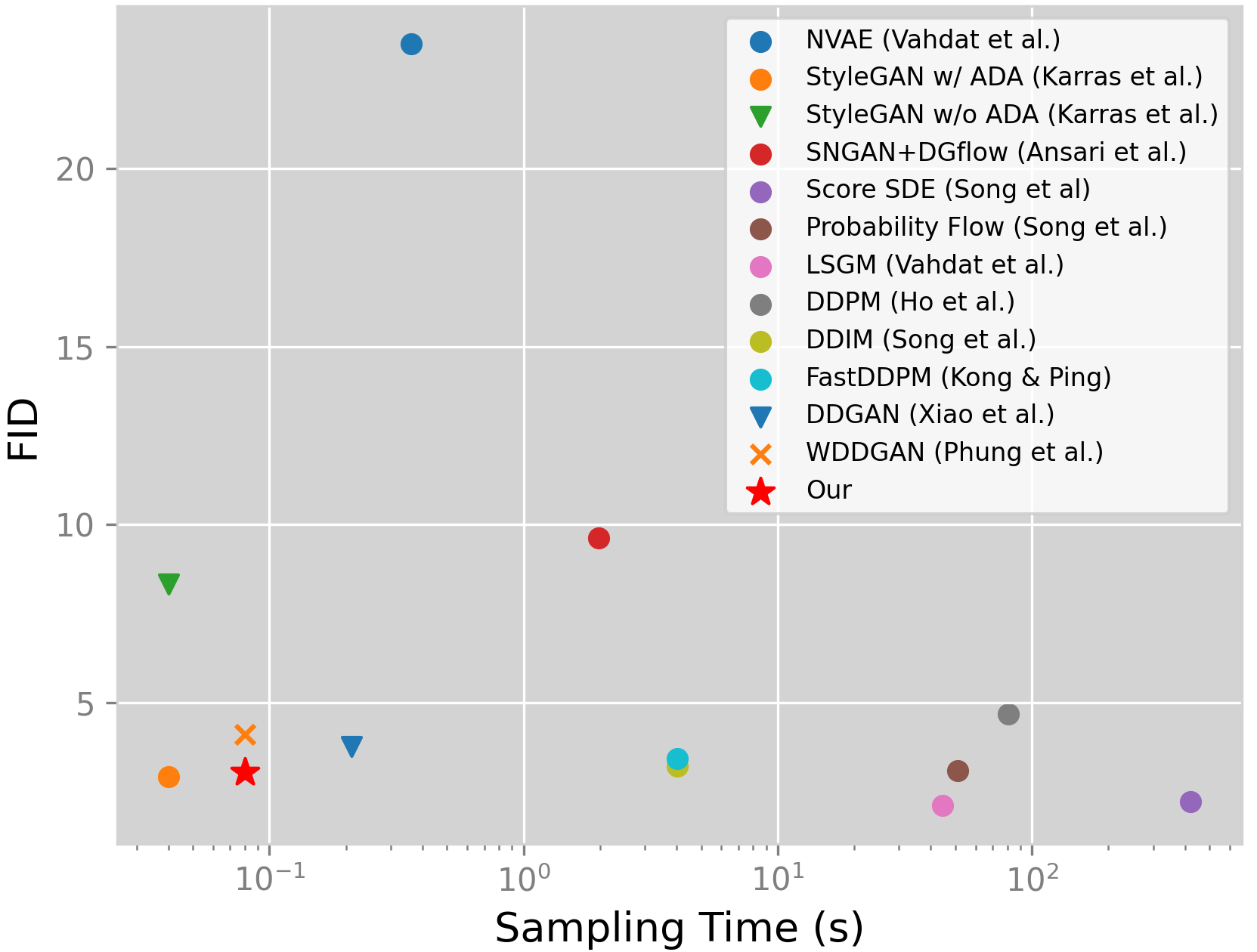}
    \caption{Sample quality vs sampling time trade-off.}
    \label{fig:tradeoff}
\end{figure}

\begin{figure}[h]
    \centering
    \includegraphics[width=0.4\textwidth]{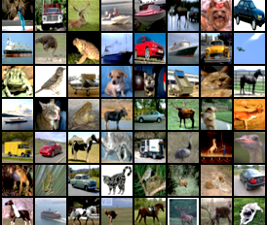}
    \caption{CIFAR-10 qualitative samples.}
    \label{fig:CIFAR10 image}
\end{figure}

\begin{table}[h!]
    \centering
    \resizebox{\columnwidth}{!}{%
     \begin{tabular}{l c c c c} 
     \hline
     Model & FID$\downarrow$ & Recall$\uparrow$ & NFE$\downarrow$ & Time(s)$\downarrow$\\ 
     \hline
     Ours & \textbf{2.98} &\textbf{0.58} & 4 & \textbf{0.08} (*) \\ 
     WDGAN \cite{wddgan} & 4.01 &0.55 & 4 & \textbf{0.08}  \\ 
     DDGAN \cite{ddgan} & 3.75 & 0.57 & 4  & 0.21\\
     \hline
     DDP \cite{DDPM} & 3.21 & 0.57 & 1000 & 80.5 \\
     NCSN \cite{NCSN} & 25.3 & - & 1000  & 107.9\\
     Adversarial DSM \cite{Adversarial DSM} & 6.10 & - & 1000 & - \\
     Likelihood SDE \cite{Likelihood SDE} &2.87& -& 1000& - \\
     Probability Flow (VP) \cite{Probability Flow} & 3.08 & 0.57 & 140 & 50.9 \\
     LSGM \cite{LSGM} & 2.10 & 0.61 & 147 & 44.5 \\
     Score SDE (VE) \cite{Score SDE} & 2.20 & 0.59 & 2000 & 423.2 \\
     Score SDE (VP) \cite{Score SDE} & 2.41 & 0.59 & 2000  & 421.5\\ 
     DDIM \cite{DDIM}& 4.67 & 0.53& 50& 4.01\\
     FastDDPM \cite{FastDDPM} & 3.41& 0.56& 50& 4.01 \\
     Recovery EBM \cite{Recovery EBM}& 9.58& -& 180&- \\
     DDPM Distillation \cite{DDPM Distillation}& 9.36& 0.51& 1&- \\
     \hline
     SNGAN+DGflow \cite{SNGAN+DGflow} & 9.62 & 0.48 & 25 & 1.98 \\
     AutoGAN \cite{AutoGAN} &12.4& 0.46 & 1 & - \\
     StyleGAN2 w/o ADA \cite{StyleGAN2} & 8.32& 0.41& 1& 0.04 \\
     StyleGAN2 w/ ADA \cite{StyleGAN2} & 2.92& 0.49& 1& 0.04\\
     StyleGAN2 w/ Diffaug \cite{StyleGAN2b} & 5.79& 0.42& 1&0.04 \\
     \hline
     Glow \cite{Glow}  & 48.9& -& 1&- \\
     PixelCNN  \cite{PixelCNN}& 65.9& -& 1024&- \\
     NVAE \cite{NVAE} & 23.5& & 0.51&0.36 \\
     VAEBM \cite{VAEBM}& 12.2& 0.53& 16&8.79 \\
     \hline
     \end{tabular}%
     }
     \caption{Results on CIFAR-10.  (*) including reference time of the decoder. Among diffusion models, our method attains a state-of-the-art speed while preserving comparable image fidelity.} 
    \label{tab:CIFAR10 results}
\end{table}
\subsubsection{Overcoming the Generative Learning Trilemma with LDDGAN}

Deep generative models face a challenge known as the generative learning trilemma, which requires them to simultaneously address three key requirements: high sample quality, mode coverage, and fast sampling \cite{ddgan}. Although diffusion models have demonstrated impressive sample quality and variety, their costly sampling process limits their practical application. DDGAN \cite{ddgan} and WDDGAN \cite{wddgan} have made significant progress in addressing the slow sampling weakness of diffusion models. The results in Table~\ref{tab:CIFAR10 results} demonstrate that our model further improves upon this weakness, achieving state-of-the-art running speed among diffusion models while maintaining competitive sample quality. While a few variants of diffusion models, such as Score SDE \cite{NCSN} or DDPM \cite{DDPM}, achieve better FID scores than our model, our model achieves a sampling speed that is 5000 times faster than Score SDE and 1000 times faster than DDPM. In particular, compared to its predecessors, DDGAN and WDDGAN, our model outperforms on all metrics considered.

We plot the FID score against sampling time of various models in Figure~\ref{fig:tradeoff} to better benchmark our method. The figure clearly highlights the advantage of our model over existing diffusion models. When comparing our model with GANs, only StyleGAN2 \cite{StyleGAN2} with adaptive data augmentation has slightly better sample quality than our model. However, Table~\ref{tab:CIFAR10 results} shows that sample diversity is still their weakness, as their recall scores are below 0.5. Figure~\ref{fig:CIFAR10 image} shows qualitative samples of CIFAR-10.

\subsubsection{High-resolution image generation}

\begin{figure*}
    \begin{subfigure}{.5\textwidth}
      \centering
      \includegraphics[width=0.98\linewidth]{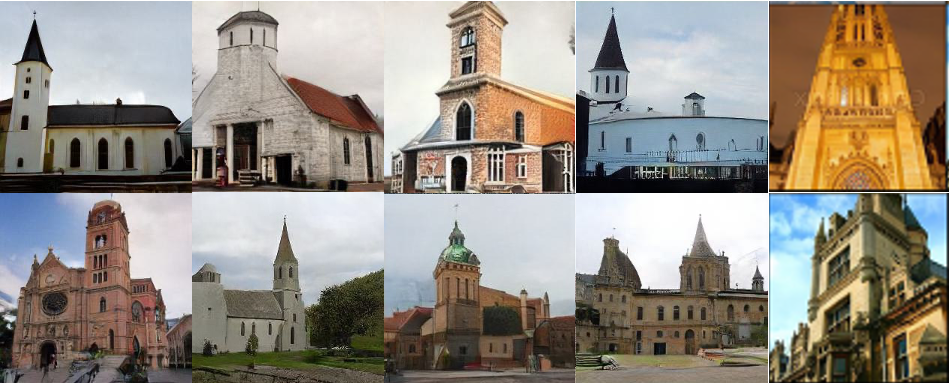}
      \caption{LSUN Church}
      \label{fig:sfig1}
    \end{subfigure}%
    \begin{subfigure}{.5\textwidth}
      \centering
      \includegraphics[width=\linewidth]{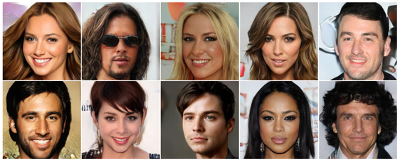}
      \caption{CELEBA-HQ}
      \label{fig:sfig2}
    \end{subfigure}
    \caption{Qualitative results on LSUN Church and CELEBA-HQ.}
    \label{fig:LSUN and CelebA}
\end{figure*}

To further evaluate the effectiveness of our model, we trained it on two high-resolution datasets: LSUN Church and CelebA-HQ. Table~\ref{tab:CelebA} and Table~\ref{tab:Lsun} present the results of comparisons with previous models. Similar to CIFAR-10, our model outperforms the two predecessors, DDGAN and WDDGAN, on all three comparison metrics. In particular, for the LSUN dataset, as shown in Table~\ref{tab: autoencoder}, successfully compressing the input data by a factor of 8 significantly reduces the inference time of the model. Our model is 1.5 times faster than WDDGAN and 3 times faster than DDGAN. When compared to other diffusion models such as DDPM or Score SDE, our model achieves comparable or better FID results. Compared to traditional GANs such as StyleGAN2, our model obtains competitive sample quality while outperforming GANs in terms of sample diversity. Figure~\ref{fig:LSUN and CelebA} displays qualitative samples from LSUN Church and CelebA-HQ.

\begin{table}[h!]
    \centering
     \begin{tabular}{l c c c c} 
     \hline
     Model & FID$\downarrow$ & Recall$\uparrow$ & Time(s)$\downarrow$\\ 
     \hline
     Ours & \textbf{5.21} &\textbf{0.40} & \textbf{0.55} (*)\\ 
     WDGAN \cite{wddgan} & 5.94 & 0.37 & 0.60\\
     DDGAN \cite{ddgan} & 7.64 & 0.36 & 1.73\\
     \hline
     Score SDE \cite{Score SDE} & 7.23 & - & - \\
     NVAE \cite{NVAE}& 29.7&-&- \\
     VAEBM \cite{VAEBM}&20.4&-&-\\
     PGGAN \cite{celeba}& 8.03 & - & - \\
     VQ-GAN \cite{vqgan}& 10.2 & - & -\\
     \hline
     \end{tabular}
     \caption{Result on CELEBA-HQ. (*) including reference time of the decoder.}
     \label{tab:CelebA}
\end{table}

\begin{table}[h!]
    \centering
     \begin{tabular}{l c c c c} 
     \hline
     Model & FID$\downarrow$ & Recall$\uparrow$ & Time(s)$\downarrow$\\ 
     \hline
     Ours & \textbf{4.67} &\textbf{0.42} & \textbf{1.02} (*)\\ 
     WDGAN \cite{wddgan} & 5.06 & 0.40 & 1.54\\
     DDGAN \cite{ddgan} & 5.25 & 0.41 & 3.42\\
     \hline
     DDPM \cite{DDPM} & 7.89 & - & - \\
     ImageBART \cite{ImageBART}& 7.32&-&- \\
     \hline
     PGGAN \cite{celeba}& 6.42 & - & - \\
     StyleGAN \cite{StyleGAN2}& 4.21 & - & -\\
     StyleGAN2 \cite{StyleGAN2b}& 3.86 & 0.36 & - \\
     \hline
     \end{tabular}
     \caption{Result on LSUN Church. (*) including reference time of the decoder.}
     \label{tab:Lsun}
\end{table}
\subsection{Ablation studies}
\label{subsec:alab}
\subsubsection{Contribution of reconstruction loss and Weighted Learning}
As described in Section~\ref{subsec: WL}, using reconstruction loss helps to achieve better FID scores but may reduce sample diversity. To leverage the effectiveness of reconstruction loss while maintaining high sample diversity, we proposed Weighted Learning, a novel method for combining reconstruction loss and adversarial loss, instead of using a simple linear combination. In this section, we investigate the impact of reconstruction loss on the training results of the model as well as the contribution of Weighted Learning. 

We trained our model using three different overall loss functions for the generator: (i) adversarial loss alone, (ii) a simple linear combination of adversarial loss and reconstruction loss similar to Eq.~\ref{eq:original loss G}, and (iii) a combination of adversarial loss and reconstruction loss using Weighted Learning. Experimental results are summarized in Table~\ref{tab: WL}.

\begin{table}[h!]
    \centering
    \resizebox{\columnwidth}{!}{%
     \begin{tabular}{c l c c} 
     \hline
     Dataset & Model & FID$\downarrow$ & Recall$\uparrow$ \\ 
     \hline
     \multirow{3}{*}{CIFAR10} &AdvLoss & 3.15 &0.58 \\ 
     &AdvLoss + RecLoss & 3.09&0.56 \\ 
     &AdvLoss + RecLoss + WL &  \textbf{3.03} & \textbf{0.58} \\
     \hline
     \multirow{3}{*}{CELEBA HQ} &AdvLoss& 5.27 &0.39 \\ 
     &AdvLoss + RecLoss & 5.23&0.36 \\ 
     &AdvLoss + RecLoss + WL & \textbf{5.21}&\textbf{0.40} \\
     \hline
     \end{tabular}%
     }
     \caption{Contribution of reconstruction loss and Weighted Learning. "AdvLoss" denotes adversarial loss, "RecLoss" denotes reconstruction loss, "WL" denotes Weighted Learning.}
     \label{tab: WL}
\end{table}

As shown in Table~\ref{tab: WL}, employing reconstruction loss yields better FID scores than relying solely on adversarial loss for both datasets. This improvement arises because reconstruction loss provides the generator with additional direct feedback regarding the disparity between the input training data and the data generated by the model, facilitating the rapid generation of data with superior image quality. However, reconstruction loss can also be interpreted as necessitating the generator to produce data identical to the input data using random noise. Consequently, after a certain amount of training, the generator tends to generate similar data with different input random noise, thereby reducing sample diversity. The results in Table~\ref{tab: WL} illustrate this reduction in recall compared to scenarios without the use of reconstruction loss.

Weighted Learning is implemented to capitalize on reconstruction loss in the early stages, gradually diminishing its priority to emphasize adversarial loss and enhance sample diversity. Towards the end of training, the rate at which reconstruction loss diminishes is reduced to ensure training stability. This strategic approach allows us to leverage reconstruction loss effectively, improving FID scores while ensuring sample diversity. As shown in Table~\ref{tab: WL}, the utilization of Weighted Learning helps us achieve better results for both FID and Recall.

\subsubsection{Affect of learned latent space by Autoencoder}
As outlined in Section~\ref{subsec:autoencoder}, introducing a KL-penalty towards a Gaussian distribution in the learned latent space proves effective for models relying on Gaussian-dependent structures (\cite{ld}, \cite{VAE1, VAE2}, \cite{ld2}). Nevertheless, such a penalty is deemed unnecessary and potentially detrimental for our model, which operates with complex and multimodal distributions. 

To verify this hypothesis, we trained autoencoders under two settings: (i) using only perceptual loss and a patch-based adversarial loss, as presented in Section~\ref{subsec:autoencoder}, and (ii) adding a KL-penalty between the learned latent and a Gaussian distribution to the aforementioned loss functions. We selected autoencoders with reconstruction FID scores close to each other to objectively compare their effectiveness through FID and Recall of the main model. 

The results in Table~\ref{tab: autoencoder alab} show that when autoencoders are allowed to freely search for the appropriate latent space, they converge significantly faster, requiring fewer epochs to reach a reconstruction FID comparable to those using a KL-penalty. When comparing the results of FID and Recall of the main model, the ability to freely search for the appropriate latent space, instead of relying on latent spaces close to a Gaussian distribution, significantly improves the results in most cases. Notably, in the case of CelebA-HQ, even though using an autoencoder with a worse reconstruction FID than one using a KL-penalty, the main model still achieves better FID and Recall scores. FID is also significantly improved in the cases of CIFAR-10 and LSUN. This result validates our hypothesis.

\begin{table*}[h!]
    \centering
     \begin{tabular}{l c c c c c c } 
     \hline
      & \multicolumn{2}{c}{Autoencoder's FID $\downarrow$} & \multicolumn{2}{c}{LDDGAN's FID $\downarrow$} &\multicolumn{2}{c}{LDDGAN's Recall $\uparrow$} \\
      \cmidrule(lr){2-3}\cmidrule(lr){4-5} \cmidrule(lr){6-7}
      & KL penalty & w/o KL penalty & KL penalty & w/o KL penalty & KL penalty & w/o KL penalty \\
     \hline
     CIFAR-10 & 1.33 (525ep)& 1.32 (300ep) & 3.15 & \textbf{3.03} & 0.58 & 0.58 \\ 
         CELEBA-HQ&  0.56 (625ep)& 0.58 (450ep)& 5.88 & \textbf{5.83} & 0.37 &\textbf{0.40}\\ 
     LSUN Church& 1.16 (675ep) &1.14 (525ep) &4.89 &\textbf{4.67} &0.41 & \textbf{0.42}\\ 
     \hline
    
     \end{tabular}
     \caption{Comparison results between using and not using a KL-penalty between the learned latent and a standard normal distribution when training autoencoders. The autoencoders are selected to have nearly identical FID reconstruction scores to ensure the objectivity of the comparison of the results of the main model. The numbers in (*) indicate the number of epochs required to train the autoencoder.}
     \label{tab: autoencoder alab}
\end{table*}
\subsubsection{The role of the autoencoder in achieving high performance}
One evident observation from the experimental results is the strong dependence of our model on pre-trained autoencoders. A proficient autoencoder plays a crucial role in compressing images to their smallest form, thereby significantly reducing computational costs and enhancing inference speed. Moreover, this compression facilitates the training of the generator by ensuring that essential features are effectively and comprehensively extracted by the encoder. The LSUN case in this study exemplifies this pattern, as demonstrated in Tables~\ref{tab: computational cost} and \ref{tab:Lsun}. Notably, well-designed autoencoders like this can also be repurposed for training future models for other tasks, such as text-to-image or super-resolution.

Conversely, our model fails to achieve optimal performance without the presence of a proficient autoencoder, as evidenced by the results on CIFAR-10 in Table~\ref{tab:CIFAR10 results}. Despite only compressing data four times, which is not significantly different from WDDGAN \cite{wddgan}, our model, by leveraging latent space features, attains superior FID and Recall scores. However, the inference speed is almost equivalent to that of WDDGAN. Additionally, in this scenario, as shown in Table~\ref{tab: computational cost}, the total number of parameters and FLOPs has increased compared to WDDGAN due to the inclusion of the autoencoder.

\section{Conclusion}
\label{sec:conclusion}
In this paper, we introduce a new diffusion model called LDDGAN. The core concept of our approach involves transitioning from the complex, high-dimensional pixel space to an efficient, low-dimensional latent space using a pretrained autoencoder. Subsequently, the GAN mechanism facilitates the diffusion process, resulting in a notable reduction in computational costs, an acceleration in inference speed, and improvements in the quality and diversity of generated images. To further enhance the quality and diversity of images, we eliminate the dependence of the latent space learned by autoencoders on Gaussian distributions and propose Weighted Learning to effectively combine reconstruction loss and adversarial loss. Our model achieves state-of-the-art inference speed for a diffusion model. In comparison to GANs, we attain comparable image generation quality and inference speed, along with higher diversity. Furthermore, when compared to two predecessors, DDGAN \cite{ddgan} and WDDGAN \cite{wddgan}, our model significantly outperforms them across the majority of comparison metrics. With these initial results, we aim to foster future studies on real-time and high-fidelity diffusion models.


\begin{appendices}
\section{Network configurations}
As mentioned in Section~\ref{subsubsec: imple}, we construct the generator based on NCSN++ \cite{NCSN} with configurations for each dataset as shown in Table~\ref{tab: generator}. The same configuration is applied for both CELEBA-HQ and LSUN. For CIFAR10, due to the relatively small input size, the model is simplified to minimize computational costs. Regarding the discriminator, we design it based on ResBlock, which is used for the generator. Details of the discriminator's architecture are presented in Tab.~\ref{tab: discriminator}. We use 3 ResBlocks for CIFAR10, 4 ResBlocks for CELEBA-HQ, and LSUN.

\begin{table}[h!]
    \centering
    \resizebox{\columnwidth}{!}{%
     \begin{tabular}{l c c c} 
     \hline
     & CIFAR10& CELEBA-HQ& LSUN \\
     \hline
     \# of ResNet blocks per scale& 2& 2&2 \\
     Base channels & 128 & 128 & 128 \\
     Channel multiplier per scale & (1,2,2) & (1,2,2,2) & (1,2,2,2) \\
     Attention resolutions & None & 16 & 16 \\
     Latent Dimension & 25 & 25 & 25 \\
     \# of latent mapping layers & 4 & 4 & 4 \\
     Latent embedding dimension & 256 & 256 & 256 \\
     \hline
    
     \end{tabular}%
     }
     \caption{Network configurations for the generator.}
     \label{tab: generator}
\end{table}

\begin{table}[h!]
    \centering
     \begin{tabular}{c} 
     \hline
     Discriminator \\
     \hline
     $ 1 \times 1$ conv2d, 128 \\
     ResBlock, 128 \\
     ResBlock down, 256 \\
     ResBlock down, 512 \\
     ResBlock down, 512 \\
     minibatch std layer \\
     Global Sum Pooling \\
     FC layer $\rightarrow$ scalar \\
     \hline
    
     \end{tabular}
     \caption{Network structures for the discriminator. The number on the right is the number of channels of Conv in each residual block.}
     \label{tab: discriminator}
\end{table}

\section{Training hyper-parameters}
Details of the hyperparameters used for the training process are presented in Table 9. It is noteworthy that we use a larger batch size than DDGAN and WDDGAN with fewer GPUs, thanks to the data compression capability of autoencoders. Regarding training times, CIFAR10 requires 1 day on 1 GPU, CELEBA-HQ requires 1.5 days on 2 GPUs, and LSUN requires 2.1 days on 2 GPUs.
\begin{table}[h!]
    \centering
     \begin{tabular}{l c c c} 
     \hline
     & CIFAR10& CELEBA-HQ& LSUN  \\
     \hline
     Learning rate of G& 1.6e-4& 2e-4& 2e-4 \\
     Learning rate of D& 1.25e-4 & 1e-4 & 1e-4 \\
     EMA & 0.9999 & 0.999 & 0.999 \\
     Batch size & 256 & 128 & 128 \\
     Lazy regularization & 15 & 10 & 10 \\
     Numner of epochs & 1700 & 400 & 400 \\
     Number of timesteps & 4 & 2 & 4 \\
     Number of GPUs&1&2&2 \\
     \hline
    
     \end{tabular}
     \caption{Training hyper-parameters.}
     \label{tab: hyper}
\end{table}

\section{Choosing backbone for latent space}
When transitioning from pixel space to latent space, we carefully consider a suitable backbone for the generator in this space. We particularly pay attention to Vision Transformers (ViTs) models due to their ability to extract long-range relationships and high compatibility with latent space. Recently, a ViT-based Unet introduced by Xiao \textit{et al.} in \cite{UViT} has demonstrated the potential to replace the widely used CNN-based Unet in diffusion models. To compare the effectiveness of ViT and CNN backbones for tasks in this study, we construct another generator based on this model. Except for the generator's architecture, we maintain the core model as depicted in Figure 1.  Timestep $t$ and $\textbf{x}_t$ are input to the generator as a token after passing through embedding layers. We introduce adaptive group normalization layers into the transformer blocks of the ViT-based Unet and control them with the latent variable $\textbf{z}$, similar to the CNN-based Unet described in Section~\ref{sec:experiments}. This helps the generator model denoising distributions within a complex and multimodal context. Other settings of the ViT-based Unet are kept consistent with [68]. We conduct experiments with three different scales of the model: Tiny, Mid, and Large, and subsequently compare the results with the CNN-based Unet.

Table.~\ref{tab:UViT compu} presents the results of the computational cost comparison, and Table.~\ref{tab:UVit on Cifar10} provides the comparison results on CIFAR10. It is evident that the ViT-based Unet achieves better results in both image quality and diversity than the CNN-based Unet. However, to outperform the CNN-based Unet, the ViT-based Unet requires more computational cost and inference time. Additionally, the CNN-based Unet converges faster, reaching its best results in only 1500 epochs, while the ViT-based Unet needs 2300 epochs. In a general comparison across five aspects: computational cost, convergence speed, inference speed, image quality, and diversity, the CNN-based Unet remains more suitable for our latent space. This result may be explained through the inherent inductive biases of CNN, which is also a weakness of ViT models. The inductive biases have helped the CNN-based Unet model features and distributions more rapidly, especially as we utilize two-dimensional latent variables, similar to the image data that CNNs are designed to process.

\begin{table}[h!]
    \centering
     \begin{tabular}{l c c c} 
     \hline
     Model & Param (M) $\downarrow$ & MEM (GB) $\downarrow$ \\ 
     \hline
     CNN based & 41.97 & 0.18 \\
     \hline
     ViT based, Tiny & 28.01 & 0.08\\
     ViT based, Mid & 43.32 & 0.21\\
     ViT based, Large & 56.17 & 0.32\\
     \hline
     \end{tabular}
     \caption{Comparison of computational cost between CNN-based and ViT-based UNet.}
     \label{tab:UViT compu}
\end{table}

\begin{table}[h!]
    \centering
     \begin{tabular}{l c c c c} 
     \hline
     Model & FID$\downarrow$ & Recall$\uparrow$ & NFE$\downarrow$ & Time(s)$\downarrow$\\ 
     \hline
     CNN based & 2.98 &0.58 & 4 & 0.08 \\ 
     \hline
     ViT based, Tiny & 4.33 & 0.54 & 4 & \textbf{0.06} \\ 
     ViT based, Mid & 3.21 & 0.57 & 4 & 0.09 \\ 
     ViT based, Large & \textbf{2.97} & \textbf{0.59} & 4 & 0.13 \\ 
     \hline
     \end{tabular}
     \caption{Comparisonal results between CNN-based and ViT-based UNet on CIFAR-10.}
     \label{tab:UVit on Cifar10}
\end{table}

\section{Additional qualitative samples and qualitative comparision}
\begin{figure}[htbp]
    \centering
    \begin{subfigure}[b]{0.4\textwidth}
        \centering
        \includegraphics[width=\textwidth]{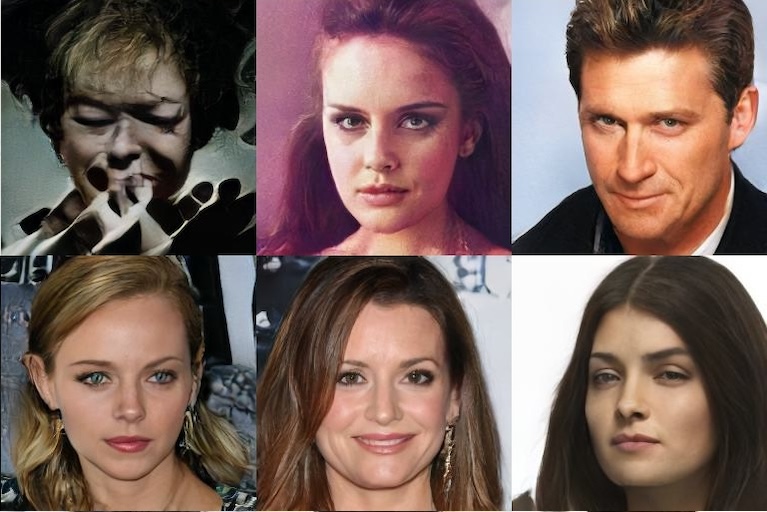} 
        \caption{DDGAN}
        \label{fig:qualitative_compare_Celeb_DDGAN}
    \end{subfigure}
    \begin{subfigure}[b]{0.4\textwidth}
        \centering
        \includegraphics[width=\textwidth]{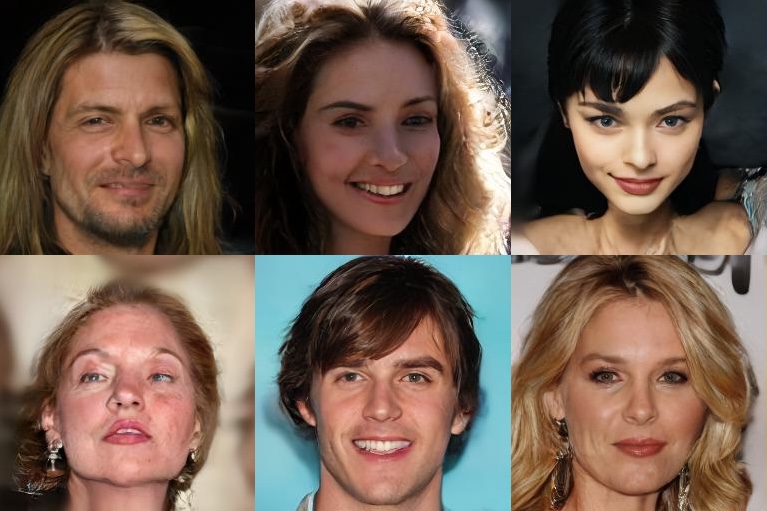} 
        \caption{WDDGAN}
        \label{fig:qualitative_compare_Celeb_WDDGAN}
    \end{subfigure}
    
    \begin{subfigure}[b]{0.4\textwidth} 
        \centering
        \includegraphics[width=\textwidth]{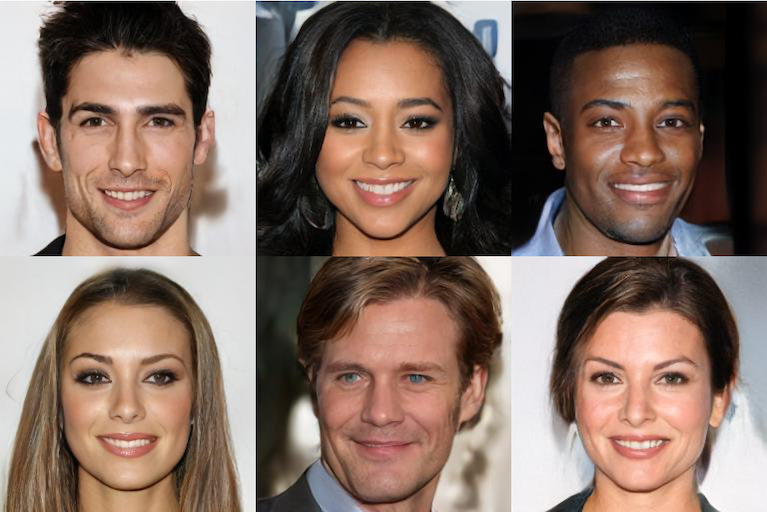} 
        \caption{Ours}
        \label{fig:qualitative_compare_Celeb_Ours}
    \end{subfigure}
    
    \caption{Additional qualitative samples and qualitative comparision on CelebA-HQ.}
    \label{fig:qualitative_compare_Celeb}
\end{figure}

\begin{figure}[htbp]
    \centering
    \begin{subfigure}[b]{0.4\textwidth}
        \centering
        \includegraphics[width=\textwidth]{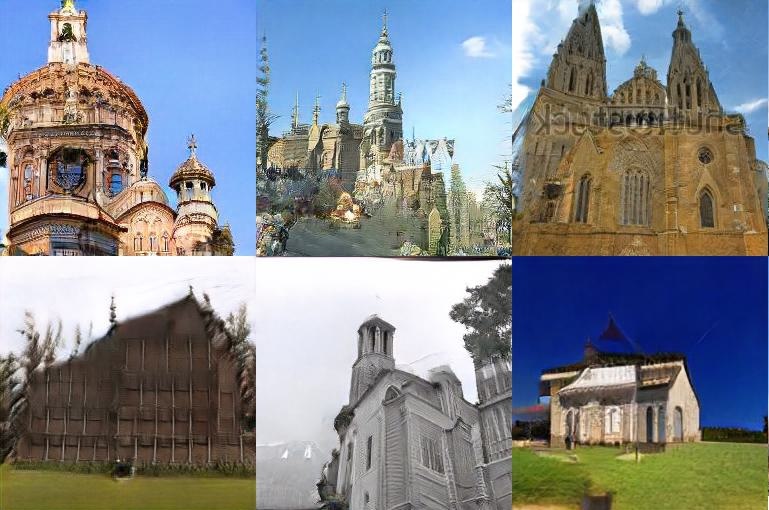} 
        \caption{DDGAN}
        \label{fig:qualitative_compare_Lsun_DDGAN}
    \end{subfigure}
    \begin{subfigure}[b]{0.4\textwidth}
        \centering
        \includegraphics[width=\textwidth]{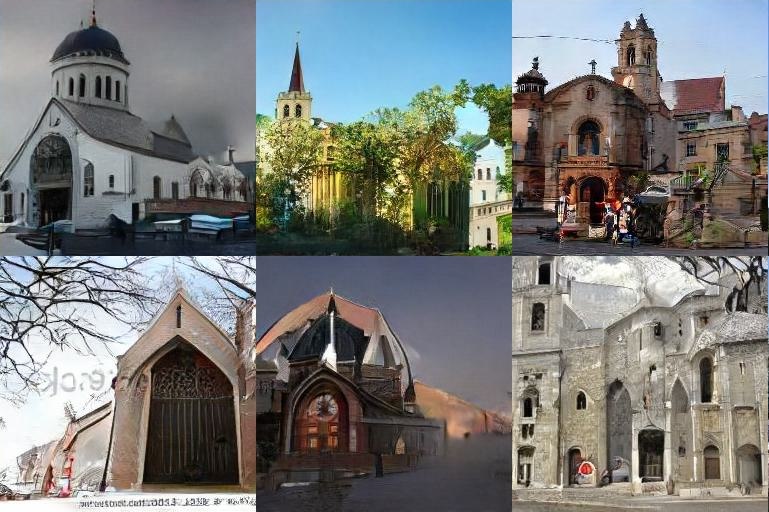} 
        \caption{WDDGAN}
        \label{fig:qualitative_compare_Lsun_WDDGAN}
    \end{subfigure}
    
    \begin{subfigure}[b]{0.4\textwidth} 
        \centering
        \includegraphics[width=\textwidth]{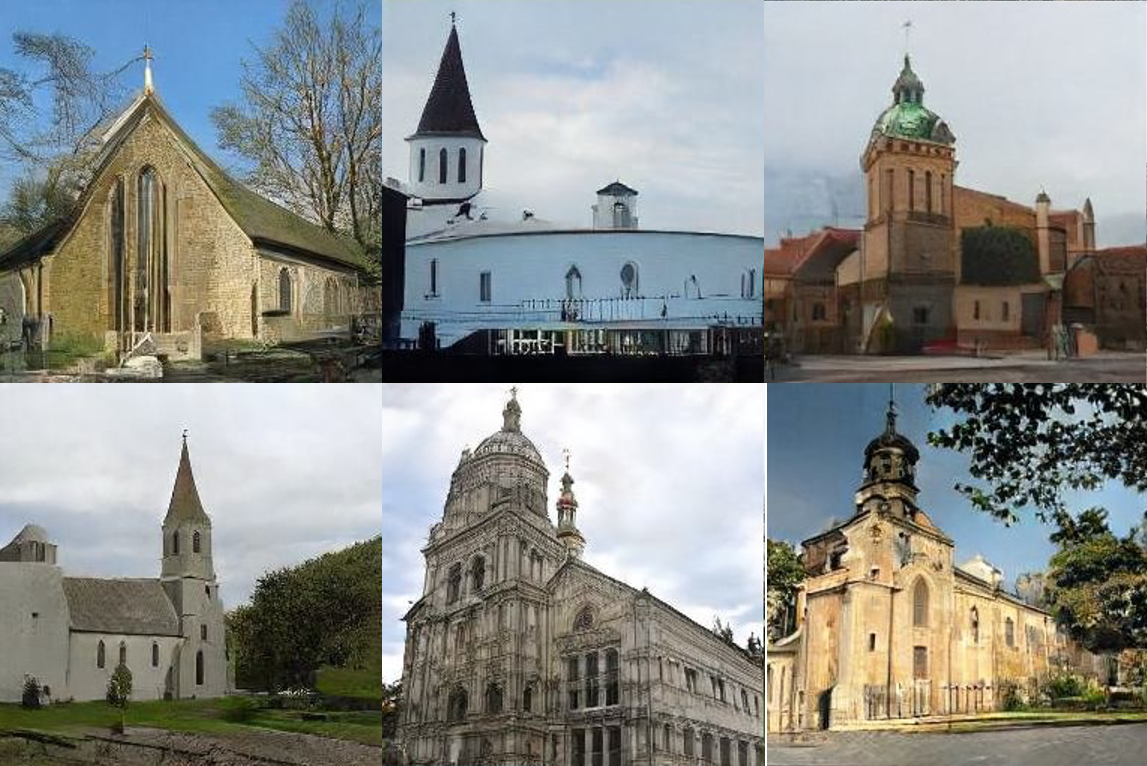} 
        \caption{Ours}
        \label{fig:qualitative_compare_Lsun_Ours}
    \end{subfigure}
    
    \caption{Additional qualitative samples and qualitative comparision on LSUN Church.}
    \label{fig:qualitative_compare_Lsun}
\end{figure}

We further provide additional qualitative samples and qualitative comparision on CelebA-HQ 256 in Figure~\ref{fig:qualitative_compare_Celeb} and LSUN-Church in Figure~\ref{fig:qualitative_compare_Lsun}.

Our model clearly achieves better sample quality. On the CELEBA-HQ dataset, both DDGAN and WDDGAN face challenges in generating distinct and complete human faces, often producing distorted features. Similarly, on the LSUN Church dataset, these models struggle to accurately render straight and horizontal architectural details. Figure~\ref{fig:qualitative_compare_Lsun_DDGAN} and ~\ref{fig:qualitative_compare_Lsun_WDDGAN} further illustrate these issues, showcasing images marred by numerous undesirable artifacts. Conversely, our model consistently produces images that are both realistic and sharply defined.

\end{appendices}
\end{document}